\documentclass[sigconf]{acmart} 
\newcommand{\modified}[1]{\textcolor{black}{#1}}

\DeclareUnicodeCharacter{0301}{*************************************}
\usepackage{microtype,stfloats,balance}
\AtBeginDocument{%
  \providecommand\BibTeX{{%
    \normalfont B\kern-0.5em{\scshape i\kern-0.25em b}\kern-0.8em\TeX}}}

\setcopyright{acmcopyright}

\copyrightyear{2023}
\acmYear{2023}
\setcopyright{acmlicensed}\acmConference[DIS '23]{Designing Interactive Systems Conference}{July 10--14, 2023}{Pittsburgh, PA, USA}
\acmBooktitle{Designing Interactive Systems Conference (DIS '23), July 10--14, 2023, Pittsburgh, PA, USA}
\acmPrice{15.00}
\acmDOI{10.1145/3563657.3595968}
\acmISBN{978-1-4503-9893-0/23/07}

\begin{document}

\title{Exploring the Design Space of \\ Extra-Linguistic Expression for Robots}

\author{Amy Koike}
\email{ekoike@wisc.edu}
\affiliation{%
  \institution{University of Wisconsin--Madison}
  \streetaddress{1210 W. Dayton St.}
  \city{Madison}
  \state{Wisconsin}
  \country{USA}
  \postcode{53706}
}

\author{Bilge Mutlu}
\email{bilge@cs.wisc.edu}
\affiliation{%
  \institution{University of Wisconsin--Madison}
  \streetaddress{1210 W. Dayton St.}
  \city{Madison}
  \state{Wisconsin}
  \country{USA}
  \postcode{53706}
  }

\renewcommand{\shortauthors}{Amy Koike and Bilge Mutlu}
\renewcommand{\shorttitle}{Exploring the Design Space of Extra-Linguistic Expression for Robots}

\begin{abstract} 
In this paper, we explore the new design space of \textit{extra-linguistic cues} inspired by graphical tropes used in graphic novels and animation to enhance the expressiveness of social robots. We identified a set of cues that can be used to generate expressions, including \textit{smoke/steam/fog}, \textit{water droplets}, and \textit{bubbles}, and prototyped devices that can generate these \textit{fluid expressions} for a robot. We conducted design sessions where eight designers explored the use and utility of these expressions in conveying the robot's internal states in various design scenarios. Our analysis of the 22 designs, the associated design justifications, and the interviews with designers revealed patterns in how each form of expression was used, how they were combined with nonverbal cues, and where the participants drew their inspiration from. These findings informed the design of an integrated module called \textit{EmoPack}, which can be used to augment the expressive capabilities of any robot platform. 
\end{abstract}



\begin{teaserfigure}
  \includegraphics[width=\textwidth]{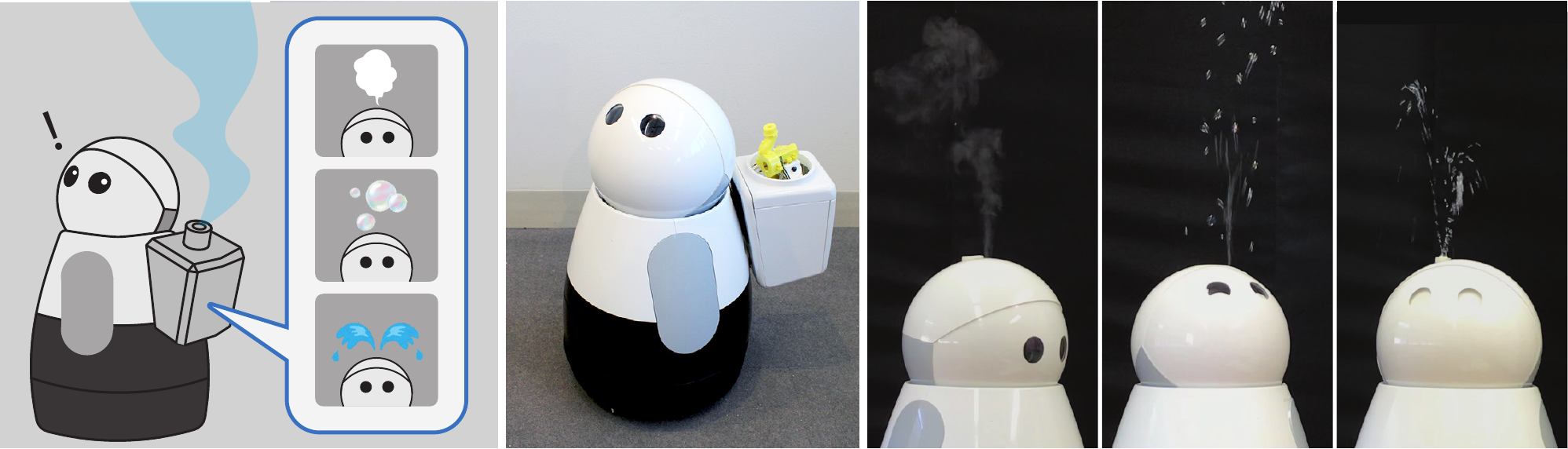}
  \caption{In this work, we explored how \textit{fluids} can be used in designing robot expressions as extra-linguistic cues. We devised mechanisms that can generate different forms of fluid effects (\textit{right}), including \textit{smoke/steam/fog}, \textit{water droplets}, and \textit{bubbles}, which can be integrated into any robot system as an augmentative add-on, which we call \textit{EmoPack} (\textit{left} and \textit{middle}). We conducted design sessions to explore how designers would use them to generate robot behaviors in various scenarios.}
  \Description[One conceptual illustration and Photos]{An illustration is on the left. A photo in the middle. A set of three photos on the right. The illustration shows a robot wearing “a backpack” that packages fluid expression devices. The photo in the middle shows an actual robot wearing “a backpack” that packages fluid expression devices, which is one of our contributions, EmoPack. Each photo shows a robot with our demonstration of fog, water droplets, and bubbles.}
  \label{fig:teaser}
  \vspace{16pt}
\end{teaserfigure}

\maketitle

\section{Introduction}

\begin{figure*}[!t]
    \centering
    \includegraphics[width=\linewidth]{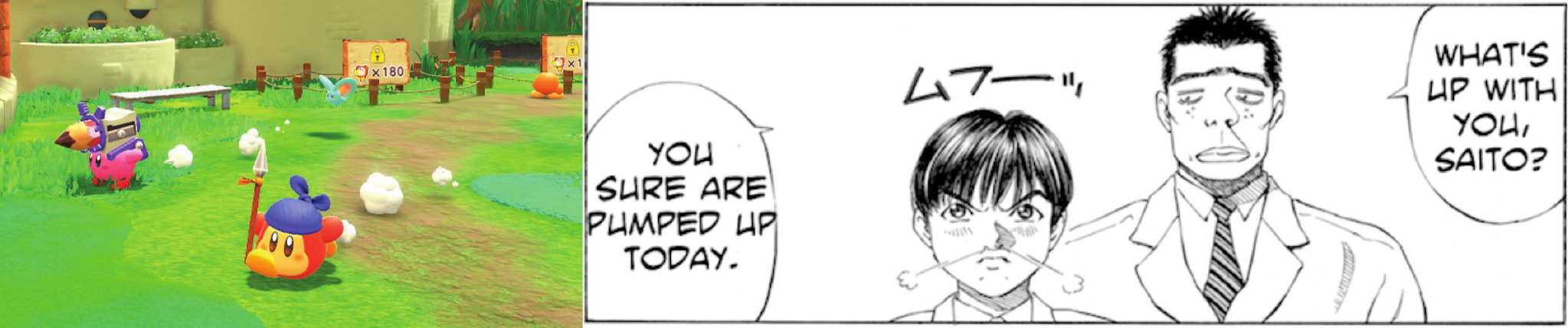}
    \caption{Examples of the use of graphic tropes in animation, video games, and comics. \textit{Left:} Smoke is drawn behind two characters, indicating the direction the character is trotting in (the image adapted from \cite{kirby}). \textit{Right:} Small puffs from the character's nose and text on his head illustrate he is pumped up (the image adapted from \cite{blackjack}). \textit{Copyright information:} left: \texttt{Nintendo Co., Ltd.}; right: \texttt{SHUHO SATO}.}
    \Description[Two illustrations]{Two images show examples of the use of Manpu and animation tropes. The image on the left shows one scene from a video game. Two characters are running in the scene, and puffs of smoke are depicted behind the characters. The other image on the left shows a panel from a comic, where a character’s actions are combined with graphical tropes.}
    \label{fig:tropeexample}
\end{figure*}

Robots designed for human interaction are also designed to use humanlike forms of communication.
These forms fall under two main modalities: \textit{linguistic} and \textit{extra-linguistic} \cite{bara1999mentalist}. Linguistic communication encompasses spoken or written language, while extra-linguistic communication refers to forms of communication that occur outside of spoken or written language, including bodily cues, visual expressions, and non-speech sounds. Research in the field of human-robot interaction (HRI) has explored how various extra-linguistic cues, such as facial expressions \cite{reyes2019robotics}, gaze behavior \cite{Multu2007}, and gestures \cite{Carter2014} can enhance human-robot interaction. 
However, what extra-linguistic cues a robot can utilize is fundamentally limited by the robot's physical design. For instance, robots that do not have movable arms to perform gestures or lack human-like features altogether cannot utilize body language. To address this challenge, previous research has explored alternative extra-linguistic cues that can augment existing physical designs and were inspired by communication modalities used in other domains. An example modality is the use of light displays as social cues, inspired by automobile signals \cite{Daniel2015LED} or bioluminescent organisms \cite{Song2018}. Another example is the design of an expandable physical system for flying robots \cite{Hooman2020}, inspired by the way pufferfish inflate their bodies when threatened or vehicle airbags that inflate during accidents, with the goal of both protecting the robot and communicating its internal states. In this paper, we add to this body of literature with an exploration of novel extra-linguistic cues.

We employ a Research through Design (RtD) approach \cite{Zimmerman2007RtDinHCI, Luria2021RtD2021} to develop new extra-linguistic cues to enhance the expressiveness of robots with limited expressivity due to their current physical designs. We found inspiration from \textit{graphical tropes} (Figure \ref{fig:tropeexample}), widely used in animation, video games, and comics to convey the character's intent and emotions. Since animated characters and robots share similar limitations in their expressive capability, graphical tropes could serve as a great resource for the design of extra-linguistic cues for robots. Through a screening process, we determined which graphical tropes can be applied to robot expression and identified three groups of graphical tropes taking the form of fluid: \textit{smoke/steam/fog}, \textit{water droplets}, and \textit{bubbles}. We explored how those \textit{fluid expressions} could be implemented and used by designers for robots through a series of design sessions. Our final design solution, \textit{EmoPack} (Figure \ref{fig:teaser}), can be attached to any robot platform and emits fluids to express the robot's internal states. 

In this paper, we present our process of designing extra-linguistic cues. In the remainder of the paper, we outline relevant work and describe our design process for incorporating expressions by manipulation of fluid as extra-linguistic cues in social robots. We present the procedures and findings of a design session that explored how designers might use fluid expressions for a robot. The paper concludes with a presentation of our final design and a discussion of our future work. Our work makes the following contributions:
\begin{enumerate}
    \item \textit{Design}: an exploration of how graphical tropes can inspire the design of extra-linguistic cues that can improve the expressiveness of robots;
    \item \textit{Empirical}: an understanding of how designers might use fluid-based expressions in designing robot behaviors; 
    \item \textit{Artifact}: \textit{EmoPack}, a backpack-style addition for robots that can be attached to any robot platform and contains a mechanism to generate fluid expressions.
\end{enumerate}

\section{Related Work} 
    Our work builds on literature on extra-linguistic communication in HRI, how animation and comics can inspire the design of robot expressions, and interaction design using fluids. We provide brief reviews of these bodies of work below. 

\subsection{Extra-linguistic Communication in HRI}
    Prior work in human-robot interaction shows that extra-linguistic cues can help people understand a robot's intent or internal states while enhancing and enriching the interaction between the human and the robot. 

\subsubsection{Bodily Cues \& Motion Design} 
    The design of a robot's motion can help communicate the robot's internal states or support people's perceptions of the robot. For example, modifying the robot's motion trajectory during its task is a way to more clearly communicate its intent and to increase its likability \cite{Dragan2015Legible, Szafir2014AFF, Knight2016, Faria2016}. Modifying the motion trajectories of the lifting action of a robot can also convey information about the object's weight \cite{Sciutti2014,praveena2020supporting}. \citet{Carter2014} studied how gestures can facilitate playing catch between a human and a robot and concluded that robot gestures can enhance physical interaction, as participants smiled more toward the robot with the gestures and rated it to be more engaging. 
    In addition to modifying motion trajectory, the design of the robot's higher-level behavior can serve as a means of extra-linguistic communication. Several studies have explored human perceptions of robot behavior designed to convey curiosity, attention-seeking, or help-seeking. \citet{Walker2020} investigated how people perceived the robots' actions during off-task and showed that such actions facilitated the perception of the robot as being curious. 
    As these studies have shown, designing bodily cues and robot motion can enhance human-robot communication, although successful use of these cues requires specific bodily capabilities or low-level manipulations of the robot's motion in ways that might be highly contextual, which may not be feasible for many robots and robot design scenarios. 

\subsubsection{Add-on Extra-linguistic Cues for Robots with Limited Expressivity}     
    Research in HRI has explored other kinds of extra-linguistic cues that add communicative capabilities to robots. These approaches include projecting information on the robot using Augmented and Mixed reality (AR/MR) and adding physical devices to robots. 

    \textit{AR/MR} approaches enable robots to enhance their communicative capacity without constraints of physical reality \cite{Suzuki2022ARRobotTaxonomy}. \citet{Walker2018} used head-worn AR to project signals to inform the motion intent of a flying robot and showed that this augmentation improved the task efficiency and user perception of a robot as a teammate. AR has also been used to add virtual arms on a robot that lacks physical expressivity to enable the use of the virtual arms in engaging with users \cite{Groechel2019}. \citet{Yound2006} proposed an AR system that projects information around a robot in the form of a thought bubble. \citet{Rosen2020} used AR to display robot motion plans and showed that it allowed users to quickly and accurately understand the robot's intent. Projection-based systems have also been proposed for human-robot collaboration \cite[\textit{e.g.},][]{Bolano2019, Andersen2016} that have been shown to help users better understand the robot's intent and perceive the robot more positively. Recent research also explored a multi-modal approach, combining haptics and AR to convey information about the robot's intent \cite{Mullen2021}.
    
    Prior work has also explored the use of \textit{physical augmentations} to enhance expressivity for robots, including several examples of light displays integrated into existing robot platforms \cite[\textit{e.g.},][]{Daniel2015LED, Cha2017, Song2018}. \citet{Daniel2015LED} investigated the use of LED strips to display light signals that communicate the directionality of motion for a flying robot. \citet{Cha2017} developed a light-signaling module that can be installed on any robot platform. \citet{Song2018} designed light signaling for a vacuum robot and evaluated its effectiveness. In addition to light, prior work explored the use of an expandable structure on a flying robot to get the attention of passersby \cite{Hooman2020}. A dog-tail attachment was used to show the robot's emotional states \cite{Singh2013Dogtail, singh2013doghci}.

    These studies point to promising design space for additive extra-linguistic cues that can be applied to various designs of robots and augment their expressive capacity.


\subsection{Animation \& Comics inspired HRI} 
\label{sec:rw:inspiration} 

    Prior research in HRI has explored how animation principles and techniques can be applied to the design of robot movement to improve communication effectiveness \cite{Schulz2019AnimationTechniqueinHRI, thomas1995illusion}. For example, \citet{Takayama2011Pixar} investigated expressions for a robot that communicated ``forethought,'' inspired by the animation techniques of anticipation and reaction. These techniques have also been applied to the motion profiles of flying \cite{szafir2014communication} and ground \cite{Schulz2019} robots to improve their favorability by nearby users. \citet{Tiago2012} applied animation principles to the design of a robot's expressions of emotion. \citet{Terzio2020} drew from character animation principles to design physical augmentation, motion trajectories, and breathing motions for a collaborative robot. Animation principles can be applied to any robot with the physical features and capability for expression, however, not all robots have enough such features or capability to design complex behavior.

    Earlier HRI research also borrowed from comics and cartoon art for augmenting robot expressivity. 
    \citet{Young2007Cartooning} explored how cartoon expression might be used with a robot using a mixed reality (MR) system that projected cartoon effects on the robot (\textit{e.g.}, adding a cartoonish face to the robot and placing icons above the robot head).
    Previous research has also explored the use of comic-style facial expressions for robots. For example, \citet{Sanda2021} applied Japanese comic art techniques (called \textit{Manpu}) of line and shadows into facial expression design for a welfare robot. \citet{Wang2021AbsractExpression} designed comic-style facial expressions for robots to investigate abstract expression for human-robot emotional interaction.
    
    These studies reveal that animation and comics can be rich resources for enhancing robot expressiveness, although the literature lacks formal and systematic explorations of how these principles and techniques might be translated into robot expressions. 

\begin{figure*}[!t]
    \centering
    \includegraphics[width=\linewidth]{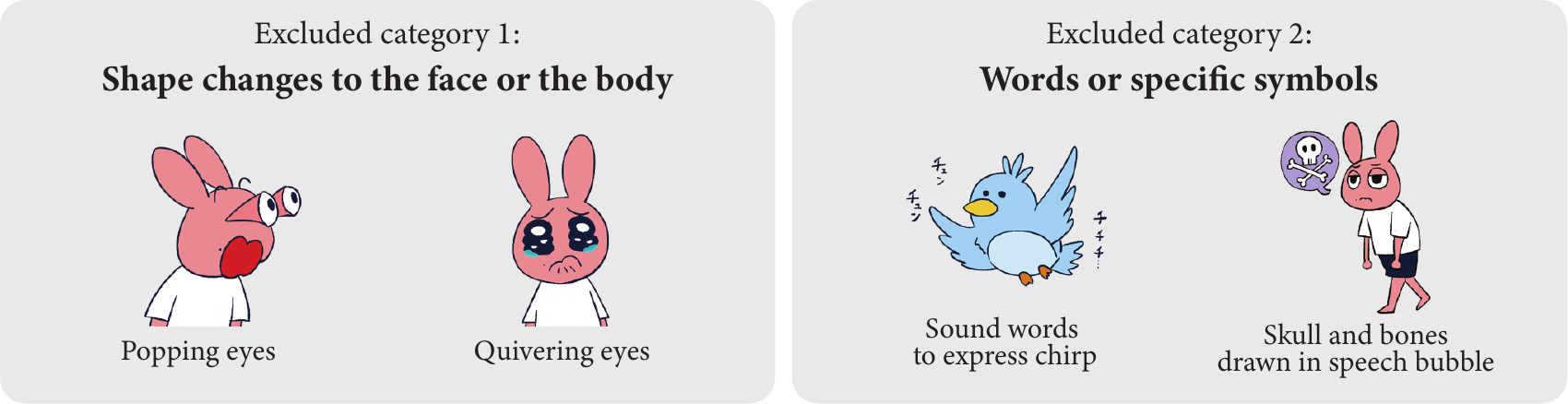}
    \caption{Examples of graphical tropes excluded in the screening process. Tropes that contain changes to the shape of the face or body of the character are not feasible for use with robots, or words and symbols that lack cross-cultural universality. See     Figure \ref{fig:group} for examples of tropes included in our design.  
    }
    \Description[Illustrations]{Four illustrations are categorized into two groups to show examples of graphical tropes that are excluded in the screening process. The first group (left) shows two graphical tropes that require shape changes on a face or body features. The second group (right) shows two graphical tropes that use words and specific symbols.}
    \label{fig:process}
\end{figure*}

\subsection{Interaction Design Using Fluids} 
    Our work also draws from the literature on the use of physical fluids in the design of tangible interfaces. Fluid-based displays can take many forms, including fog, streams or drops of water, bubbles, and olfactory stimulation. Prior work has explored the use of interactive fog displays \cite{Lam2015, Tokuda2017} and displays made from streams of water \cite{heiner1999information, Nakagaki2016}. Several projects developed transitory displays using fog, liquid, and/or soap bubbles \cite{Sylvester2010, Suzuki2017, Alakarppa2017}. In addition to the use of fluids in interactive displays, \citet{Yoshida2021} proposed a wearable device that generated pseudo tears. \citet{Lei2022} developed a toolkit for olfactory stimulation and explore its use.
    
    Research in HRI also includes a few early examples of the use of fluids for expression. \citet{Guo2020} proposed a sweating robot that generates water drops from its body to simulate sweat in order to communicate to its users that the temperature in the environment is high. In an exploratory project, \citet{Lee2019} designed a robot that generated bubbles and explored interaction possibilities with the bubble-generating robot in a public space. 
    Although these projects serve as early examples of the promise of the use of fluids in expressive HRI design, the literature lacks a comprehensive exploration of the design space of such cues, their use in communicating a range of internal states, and investigations of how they might be used by interaction designers.


\subsection{Research Questions}
    Our review of the body of work discussed above motivated three key research questions surrounding the design space of new extra-linguistic cues for such robots:
    
    \textit{RQ1: What metaphors might serve as effective extra-linguistic cues for social robots?}
    
    \textit{RQ2: How can we implement extra-linguistic cues for robots with limited expressive capability?}
    
    \textit{RQ3: How do designers use the extra-linguistic cues to design robot expressions?}
    
    \noindent Following a Research through Design (RtD) approach \cite[]{Zimmerman2007RtDinHCI, Luria2021RtD2021}, we investigated these questions by prototyping systems to generate fluid expressions, conducting design sessions to understand how they might be used, and developing the updated, final design based on our findings from these sessions.

\section{Design Process}\label{sec:designprocess}

In this section, we answer the first research question, \textit{RQ1: What metaphors might serve as effective extra-linguistic cues for social robots}, by describing a process for examining new social cues for robots. 
As previous research has shown, extra-linguistic cues can draw inspiration from a range of sources such as animals or industrial products \cite[\textit{e.g.},][]{Hooman2020, Daniel2015LED}. Choosing from many possibilities for inspiration, we focus on \textit{graphical tropes} used for animation and comics because prior HRI research (see \S\ref{sec:rw:inspiration}) has built foundational support for designing effective communication modalities based on animation and comic art techniques. Graphical tropes are visual elements that help create dynamic expressions and storylines. They use abstract visual language to convey information and enhance storytelling capability. By applying this concept to the development of extra-linguistic cues for robots, we aim to create more engaging and expressive human-robot interactions.

\subsection{Design Inspiration} 
We found inspiration for new extra-linguistic cues in graphical tropes which are used for animation, video games, and comics to enhance characters' expressivities. The graphical tropes are typically drawn on the characters' bodies or surroundings to convey the characters' emotions and actions visually (Figure \ref{fig:tropeexample}). In Japanese comics, the graphical tropes are sometimes called \textit{Manpu}, which is a made-up word that combines ``Manga'' which means comics, and ``Fugou'' which means symbols or icons. 
Although there is a slight difference between graphical tropes in animation and \textit{Manpu} (\textit{i.e.}, animations are dynamic, and comics are static), they share common expressions to communicate the same meaning. 
Using graphical tropes has several benefits: they can help the audience quickly understand the situation, exaggerate the emotions being portrayed, and add humor to a scene, despite the constraints of visual media such as a limited frame rate, static medium, limited color palette, and a limited number of pages. These tropes help to make visual media more understandable, engaging, and compelling.

Robots, like visual media, have limitations in their ability to express themselves. These limitations can include the design metaphors they follow, their mechanical structure, and their range of movement. For example, the Kuri robot, the robot platform we used in our design sessions (\S\ref{sec:designsession}), has an abstract humanoid appearance with non-movable arms.
Just as graphical tropes can be used to overcome the limitations of visual media and make them more expressive, we believe that they can augment robots' expressive capabilities.  Our approach is similar to earlier work by \citet{Young2007Cartooning} that explored the use of cartoon expressions with a robot. They argued \textit{``cartoon art [...] allows a robot to break free from the limitations of its physical body and gesture capabilities, freely adding colour, animation, and annotations at any location on or around its body.''} 


\begin{figure*}[!t]
    \centering
    \includegraphics[width=\linewidth]{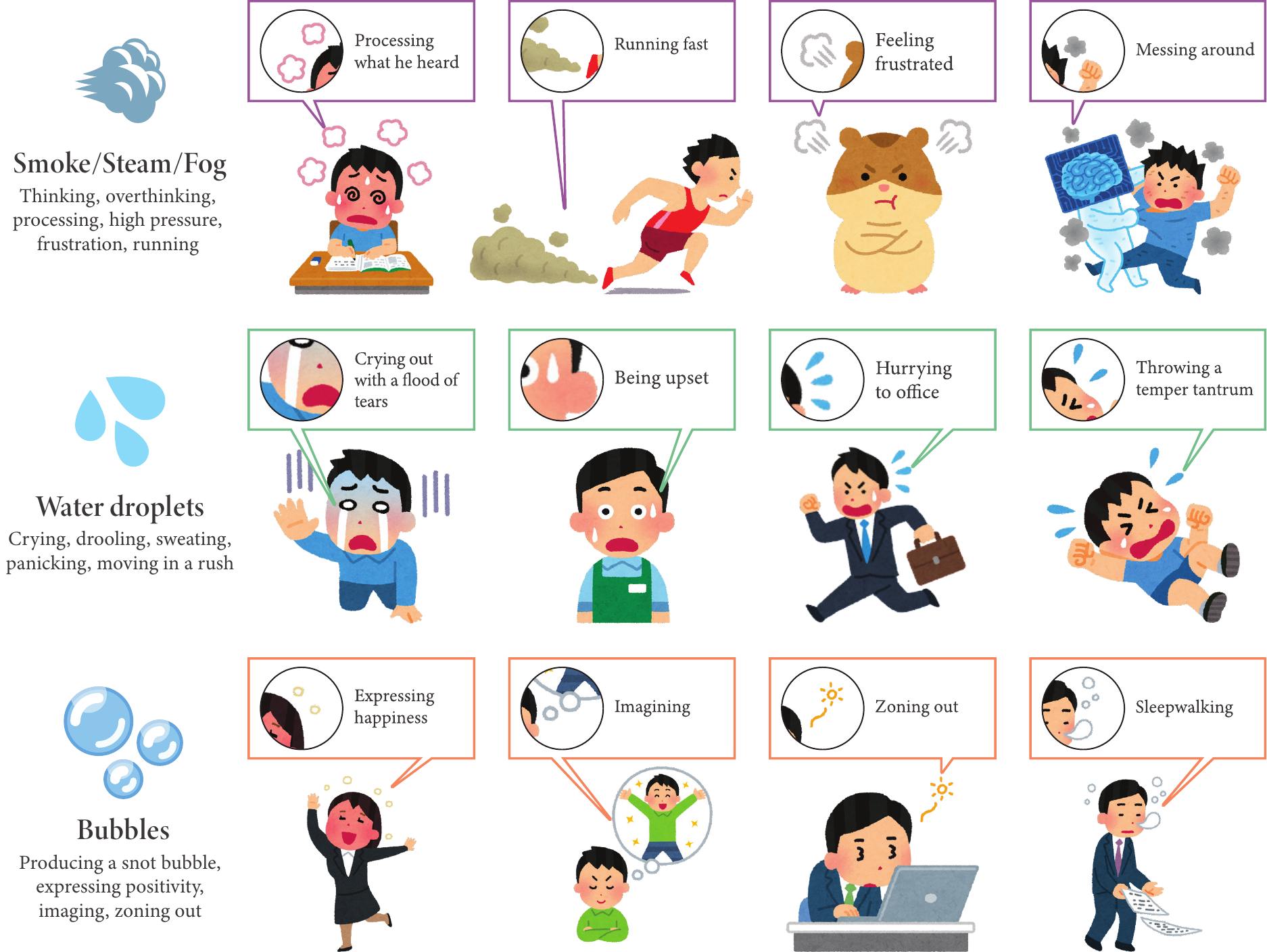}
    \caption{Three groups of graphical tropes included in our design, which took the form of fluids: (1) smoke/steam/fog, (2) water droplets, and (3) bubbles. These three groups are widely used in animation, video games, and comics to express characters' emotions, signal future action, or convey the mood of a scene in various contexts. Each image shows how the graphical trope is depicted and what it is communicating. \textit{Copyright information:} \texttt{Irasutoya}. 
    }
    \Description[Annotated illustrations in a diagram]{Twelve illustrations are categorized into three groups. The first group(top) shows four illustrations which include characters expressing processing, running fast speed, frustration, and messing around through steam-shaped graphical tropes. The second group(middle) shows illustrations where a character expresses crying, upsetting, hurrying, or desperately asking for a favor through a droplet-shaped graphical trope. The last group shows four illustrations in which characters express being in pleasant moments, expanding their imagination, thoughtfulness, and falling asleep through bubble-shaped graphical tropes.}
    \label{fig:group}
\end{figure*}

\begin{figure*}[!t]
    \centering
    \includegraphics[width=\linewidth]{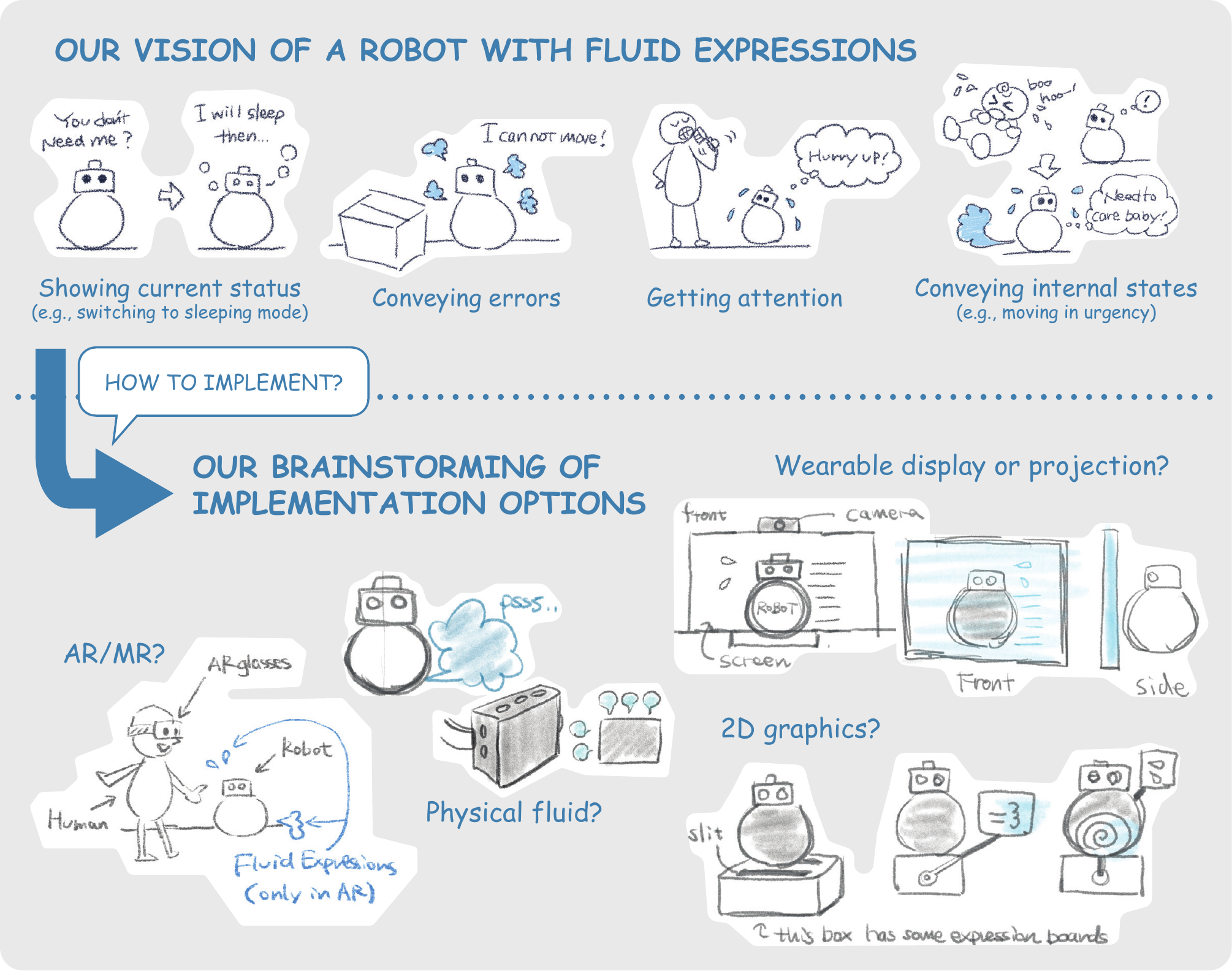}
    \caption{Annotated sketches from brainstorming how robots might use the included graphical tropes for communication (\textit{top}) and of how they might be implemented (\textit{bottom}). Among several possibilities, including the use of AR/MR, robot-wearable displays, and 2D graphics displayed on movable boards, we pursued the use of physical fluid for expression.}
    \Description[Annotated sketches]{Four illustrations are aligned on the top, and four other illustrations are placed on the bottom. All of them are hand-written illustrations. The illustrations on the top are showing the visions of a robot with fluid expressions. The illustrations on the bottom show implementation ideas.}
    \label{fig:concept}
\end{figure*}

\subsection{Design Identification \& Screening} 
Once we recognized graphical tropes as potential resources for extra-linguistic cues, we further investigated the variation of graphical tropes and studied how they are used. We examined over 200 kinds of graphical tropes introduced by \citet{manpuzufu} and in the \citet{animetrope} online repository and used a systematic screening process to select candidates that can serve as new extra-linguistic cues for robots. Our process can be broken down into two steps. First, we screened graphical tropes to determine which tropes could be included or should be excluded for further analysis. Second, we analyzed the candidates by categorizing them into archetypes based on visual similarity. 

Our screening process integrated two viewpoints. The first viewpoint was the Manpu classification done by \citet{manpuzufu}. Graphical tropes were classified into three groups in terms of \textit{how/where it is drawn}: (a) drawn apart from a character; (b) drawn to be attached or generated from an object or a character; and (c) drawn as a part of the action or body. Graphical tropes used in animation and video games also can be categorized in the same way. We excluded the type of graphical tropes classified in group (c), as it requires changes in the face or body of a character. Because we envision new extra-linguistic cues for robots with limited expressive capacities (\textit{e.g.}, a robot with no face, or a robot with no arms), we included tropes that overlap with a character or its surroundings in groups (a) and (b). 
The second viewpoint captured \textit{what it includes/what it looks like}. Specifically, we removed tropes that included words. In comics, graphical tropes often contain \textit{onomatopoeia}, words that imitate or suggest the source of the sound. Onomatopoeia or sound words, however, are not universal, meaning that different languages have different onomatopoeic words to describe the same sounds (such as descriptions of the barking of a dog as ``woof, woof'' in English, ``wan, wan'' in Japanese, and ``hav, hav'' in Turkish); thus, we excluded the word-style graphical tropes. In addition to the word style tropes, we left off graphical tropes that illustrate literal figures such as a ribbon, skull, bomb, or an angel halo. Such graphical tropes are often used to show characters' personalities or specific contexts (\textit{e.g.}, an angel halo is used to indicate an angelic personality). We employed graphical tropes that could be applied to various contexts; thus, we excluded graphical tropes with specific illustrations. Several examples of the excluded graphical tropes are shown in Figure \ref{fig:process}.

In the remainder of our screening process, we first grouped the remaining graphical tropes by visual similarity. Through discussions, we found three groups of expressions that involve forms of fluid: (1) \textit{smoke, steam, or fog}, (2) \textit{water}, and (3) \textit{bubbles}. These tropes are widely used in both comics and animations to express character emotion, signal future action, or convey the mood of a scene in various contexts. 
After we identified our focus on the three categories of expressions, we sought to explore their potential for communicating the internal states of a robot. To this end, we reviewed media where these expressions were used and extracted characters' internal states, emotions, and other contextual information conveyed in these stories. For example, we observed that drops or squirts of water were commonly used to express crying, drooling, or sweating due to panic, rushing, or stress. Figure \ref{fig:group} provides examples of these expressions and the potential uses we identified for each category. These mappings were further explored in our design sessions, which we present in \S\ref{sec:designsession}.






\subsection{Our Vision of the Use of Fluid Expression} 
Because all chosen expressions involved fluids in some form, we call them ``fluid expressions.''  Our vision of using fluid expressions for the robot involves (1) \textit{smoke, steam, or fog} exuding from the robot's head or torso; (2) \textit{water} spraying upward or in multiple directions from the robot's head or torso; and (3) \textit{bubbles} forming and flowing upward from the robot's head or torso. For example, steam exuding from the robot's head can show that the robot is processing what it received as input. The robot might express ``crying'' by producing drooling water to express the need for attention or help. Producing bubbles could show that the robot is ``dreaming'' in sleep mode. We generated such ideas for the design space of fluid expression to convey the internal and affective states of a robot, as illustrated in Figure \ref{fig:concept}. 

\section{Implementation} \label{sec:imp}

\begin{figure*}[!t]
    \includegraphics[width=\linewidth]{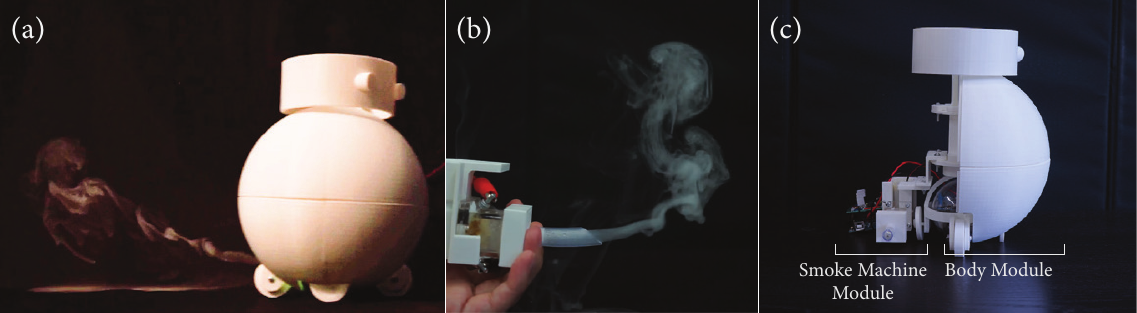}
    \caption{(a) Our earlier prototype of fluid-based expression involved (b) a fog-generation device (c) integrated into the body of a bespoke mobile robot, which informed the design of the fluid mechanisms presented in this paper.
    }
    \Description[Annotated photos]{Three photos labeled (a), (b), and (c) show our earlier prototype of fluid based expression robot. Image(a) shows a spherical-shaped robot with a smoke machine moving across a stream of smoke. Image (b) shows the smoke machine module. Image (c) shows how the smoke machine module and the body module are integrated into a bespoke robot.}
    \label{fig:early}
\end{figure*}

\begin{figure*}[!t]
    \includegraphics[width=\linewidth]{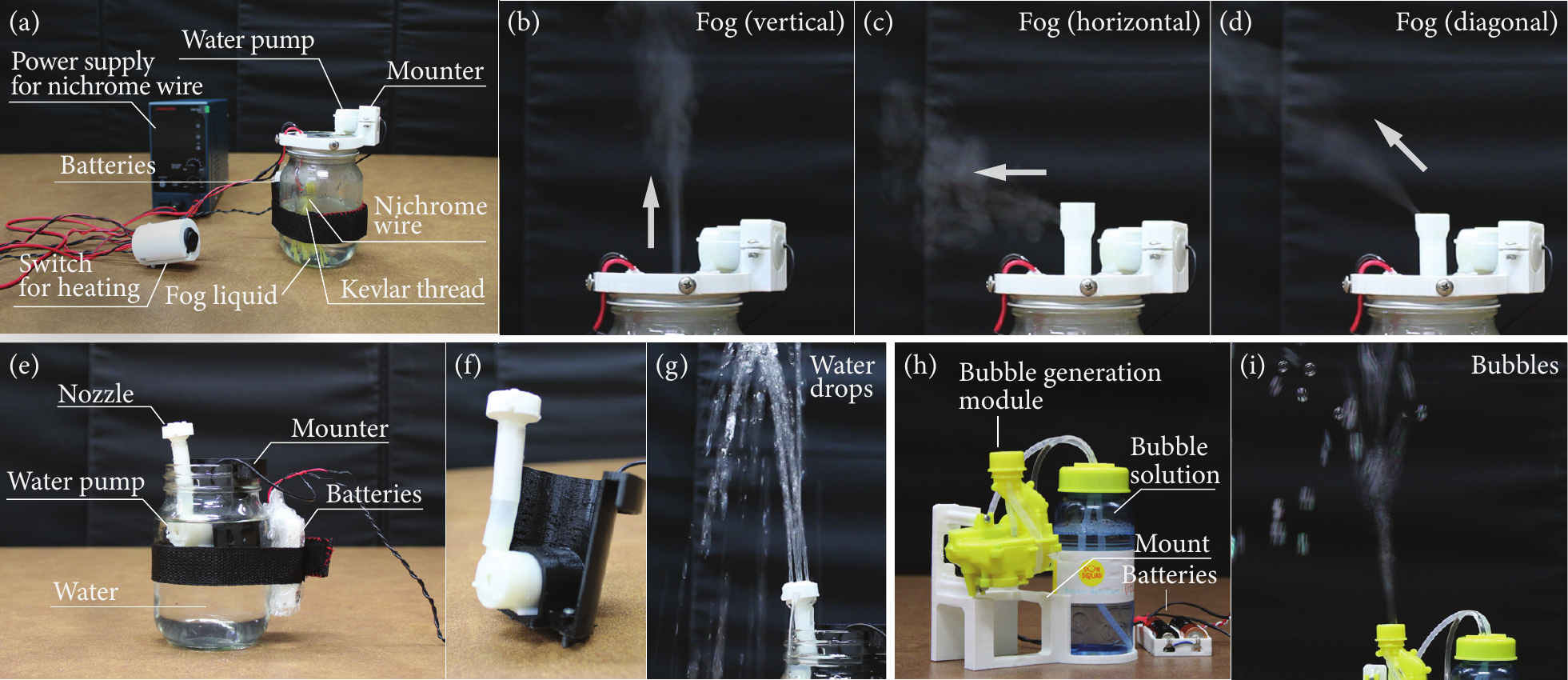}
    \caption{Prototype details for our fluid expression devices to generate (a--d) fog, (e--g) water droplets, and (h--i) bubbles.}
    \Description[Annotated photos]{Nine photos labeled (a) to (i) show the fluid expression devices’ configuration and demonstration. Images (a),(e),(f), and (h) show how components such as the water pump, mounter, nichrome wire, kevlar thread, fog liquid, switch for heating, batteries, and power supply are integrated into the fluid expression device. (b)-(d) demonstrate three patterns of smoke; in (b) fog goes vertical, in (c) fog goes horizontal, and in (d) fog goes horizontal.}
    \label{fig:implementation}
\end{figure*}

In this section, we address the second research question, \textit{RQ2: How can we implement the extra-linguistic cues for robots with limited expressive capability}, by discussing how we explored the implementation of the fluid expressions identified in \S\ref{sec:designprocess}. To implement the fluid expressions (smoke/steam/fog, water, bubbles) into robots, we selected fluids as a physical form for our implementation among several possibilities in order to realize an outstanding communication channel for robots. Following the initial prototyping (Figure \ref{fig:early}), we developed a set of fluid expression devices for further design exploration in \S\ref{sec:designsession}.\footnote{Details of our implementation; materials and procedures for the design sessions; and data from the design sessions, including the videos of designed expressions, can be found at the following \textit{Open Science Foundation (OSF)} repository: \url{https://osf.io/uwjz8/}}

\subsection{Exploration of Implementation Options}
After we identified the three groups of extra-linguistic cues, we explored how these cues might be implemented and integrated into a robot. Through brainstorming, we considered implementing them as two-dimensional graphics that are installed on movable plates that extend from the robot, projection into surfaces behind or beneath the robot, augmented-reality mapped on the environment, seen on a robot-wearable display, and effects that are generated out of fluids (Figure \ref{fig:concept}).

Among the possibilities listed above, we ultimately decided to explore the use of physical fluid for several reasons. First, we were inspired by the promise of fluid-based effects that were explored in prior research \cite[\textit{e.g.},][]{Yoshida2021, Guo2020}. We were particularly interested in exploring what fluid effects that were generated differently would convey. For example, rapidly splashing water up vs. slowly drooling water down might communicate different states. This facet of a fluid-based approach also better reflected the rich expressions offered by graphical tropes.
Second, the prevalent use of fluids in stage design inspired us to explore their use for HRI design, especially given prior explorations of HRI as theatre \cite[\textit{e.g.},][]{lu2011human,lemaignan2012roboscopie}.
Finally, because robots are embodied agents, we believe that physical fluids can serve as a better match to the embodiment of the robot, enabling the robot to generate more believable, noticeable, and evocative \textit{in situ} expressions.

\subsection{Design of Fluid Expression Devices}
To explore the design space and feasibility of the use of actual fluid for robots, we took the following steps: (1) initial prototyping;  (2) developing on-demand fluid expression devices; (3) conducting design sessions with the devices; (4) finalizing the design. In the remainder of this section, we describe the initial implementation detail (steps 1 and 2). The design session (step 3) is described in \S\ref{sec:designsession}, and the final design (step 4) is presented in \S\ref{sec:discussion:emopack}.

Our initial prototype integrated fluid expression devices within a bespoke, 3D-printed mobile social robot is shown in Figure~\ref{fig:early}. We derived several design challenges for fluid expression devices from the prototype to be reflected in the next development. For example, fluid might run out after a few hours of operation; thus the system configuration has to allow the users to refill the fluid easily. 

After the initial prototyping, we developed a set of devices to generate three expressions (smoke/steam/fog, water droplets, bubbles) on demand to conduct design sessions with designers in order to explore their use and perceptions of the devices and the expressions. The fluid expression devices are shown in Figure \ref{fig:implementation}, and each device generates either fog, water droplets, or bubbles. Devices can be switched \texttt{ON} and \texttt{OFF} using a physical button and will continue generating their fluid expression as long as the button is pushed. All devices were battery-operated to minimize cabling and afford mobility. Each device is described in further detail below.

\textit{Fog ---} To produce fog, we developed a small machine that generates directed fog on demand (Figure \ref{fig:implementation} (a--d)). The fog is generated by heating tri-ethylene glycol, commonly called ``fog liquid'' or ``fog juice.'' The machine consists of a glass jar filled with fog liquid, a kevlar thread, a nichrome wire, and a water pump. The nichrome wire is wrapped around the kevlar thread and heats the fog liquid. The pump is assembled to generate airflow inside the jar by using a 3D-printed mount to push the stream of fog out of the jar. We also designed nozzles that designers can switch to change the direction of the generated fog during the design session.

\textit{Water Droplets ---} To produce water droplets, we developed a device that generates bursts of water using a jar filled with water, a water pump, a 3D-printed mount, and a nozzle (Figure \ref{fig:implementation} (e--g)). The pump was submerged in the water using a 3D-printed mount and pushed water through the nozzle to generate splashes.

\textit{Bubbles ---} To produce bubbles on demand, we used a bubble generation module from a bubble wand toy, which we placed within a custom 3D-printed enclosure (Figure \ref{fig:implementation} (h) and (i)).

\begin{table*}[!b]
    \caption{The scenarios for which designers were asked to create robot expressions during design sessions.}
    \Description[Table]{The list of scenarios is shown in the table with six rows and four columns.} 
    \label{table:scenario}
    \centering
    \small
    \begin{tabular}{p{0.05\linewidth} p{0.1\linewidth} p{0.38\linewidth} p{0.38\linewidth}}
    \toprule
    \textbf{\textit{ID}} & \textbf{\textit{Role}} & \textbf{Context} & \textbf{Internal State} \\
    \toprule
    \textbf{D} & \textit {Delivery} & The robot is making a delivery, but something prevents the robot from moving forward. & The robot is frustrated about being stuck. \\
    \midrule
    \textbf{E} & \textit{Education Assistant} & The robot is teaching children concepts in a science class or at a science museum. & The robot is expressing curiosity to elicit interest in children. \\
    \midrule
    \textbf{P} & \textit{Patrol} & The robot detects something unusual that it must inspect. & The robot needs to rush to the location. \\
    \midrule
    \textbf{G} & \textit{Guide} & The robot is giving people a tour, but the group unexpectedly encounters construction. & The robot is processing to find a new route. \\
    \midrule
    \textbf{K} & \textit{Playing with children} & When the robot is playing with children, one of them kicks the robot. & The robot tries to discourage children from hitting it by showing that it feels hurt. \\
    \bottomrule
    \end{tabular}
\end{table*}

\section{Exploring Use of Fluid Expression} \label{sec:designsession}


This section addresses the third research question, \textit{RQ3: How do designers use the extra-linguistic cues for robots' expression}, by describing a design session that consisted of individual design activity using the fluid expression devices developed in \S\ref{sec:imp} and the results. The goal of each design session was to explore the use of fluid expressions including design patterns, draw designers' perceptions towards fluid expressions for robots, and gain feedback on practical and technical challenges such as manipulating fluids. 
In this section, we first describe the study design, then discuss the findings.

\subsection{Study Design} 
Each design session \modified{involved one participant and} lasted approximately 60 minutes. The main activity in the session was designing expressions for a set of scenarios using a robot and our prototype fluid-generation devices. 
We recorded the expressions as video footage, which we later utilized for our interviews and analysis. Participants were asked to take a video of their designs after completing each scenario. After taking a video, we conducted a semi-structured interview to ask about their design justification. This design-finalize-interview process was repeated for as many scenarios as time permitted. After the design activity, we asked participants to reflect on their experiences with fluids in a 10-minute interview. 
We describe our study design in further detail below.

\begin{figure*}[!t]
    \includegraphics[width=\linewidth]{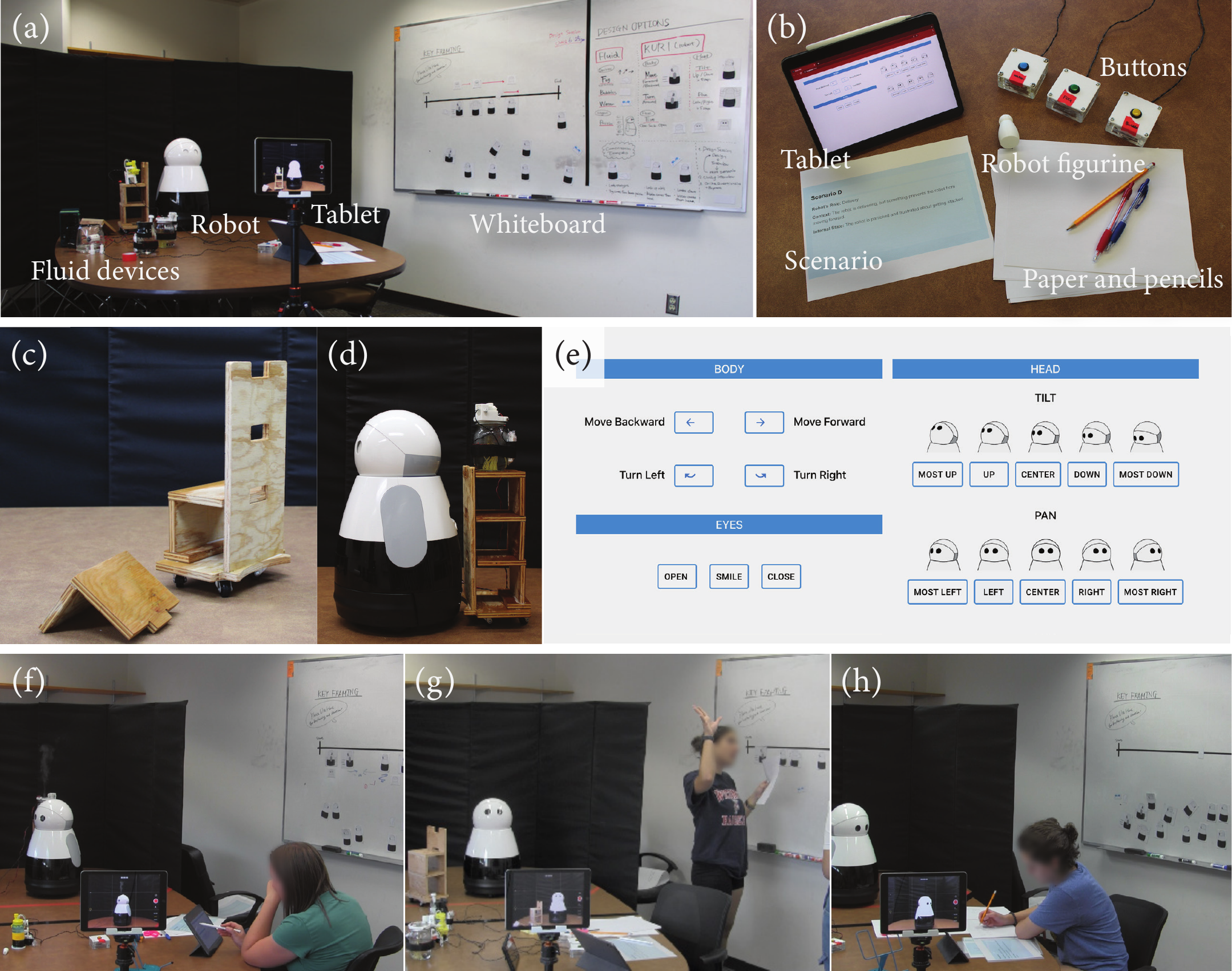}
    \caption{
    \textit{Top:} The setup of our design session. (a) The three fluid devices, the robot, and a tablet with the robot control interface were placed on a table; a second tablet was used to capture the final design; and the whiteboard facilitated design ideation and planning. (b) The design materials provided to each designer: the tablet to control the robot, buttons to control the fluid devices, scenarios they were asked to design for, and supplies and props to facilitate design ideation.
    \textit{Middle:} For the design session, we created (c--d) a mobile shelf system that could be reconfigured and moved along with the robot to support mobility during expression in order to simulate fluid expressions installed in the robot. (e) The control interface we built for the Kuri robot.
    \textit{Bottom:} Scenes from design sessions: (f) designer exploring how fog and robot motion might be combined, (g) designer using the whiteboard and body-storming expressions for the robot, and (h) designer sketching ideas for brainstorming.
    }
    \Description[Annotated photos]{On the top, two photos are showing the design session setup and are labeled (a) and (b). Photo (a) shows the room setting which has a table and a whiteboard. On the table, there is the robot, fluid devices, a tablet, and other creative resources. Photo (b) demonstrates the materials provided to the designer, such as a tablet, scenario paper, mini figure, paper and pencils, and buttons for fluid devices. In the middle, three photos inform the study materials related to the robot and are labeled (c), (d), and (e). Photos (c) and (d) shows a wagon to carry the fluid devices. Photo (e) shows the web-based interface to control the robot; in (e), there is a GUI for moving the robot’s body and eye states. On the bottom, three photos are showing scenes from the design session and are labeled (f), (g), and (h). In photo (f), a participant controls the robot and sees the fog expression. In the photo (g), a participant is standing in front of the whiteboard. In photo (h), a participant is sitting and drawing on paper for idea generation}
    \label{fig:setting}
\end{figure*}

\subsubsection{Design Tools}
For the design sessions, we used the set of fluid expression devices described in \S\ref{sec:imp} and a social robot designed for the home environment, named Kuri, which has been used in prior HRI research \cite[\textit{e.g.},][]{Groechel2019, Walker2020, hu2021exploring}. For the purposes of this study, we developed a web-based interface to allow participants to directly control the robot's movements (Figure \ref{fig:setting} (e)). Participants can use Kuri's capabilities of moving forward/backward; turning left/right at a slow speed; tilting/panning head in five steps; and displaying three eye states (open, close, and smile).
In addition to the fluid expression devices and the robot, we provided a set of cards that can be placed on a magnetic whiteboard depicting each design option as well as creative tools such as pencils, paper, and a mini figure of the robot to facilitate idea generation. These materials and study settings are shown in Figure \ref{fig:setting} (a--b).

\subsubsection{Scenario Development}
We developed five scenarios for the design sessions that are shown in Table~\ref{table:scenario}. In the development of these scenarios, we envisioned common applications where social robots are currently used or envisioned for future use and identified communication needs that would require the robot to convey its internal states. Each scenario listed a role or task for the robot (\textit{i.e.}, delivery, education assistant, guidance, patrol, playing with kids for Scenarios D, E, G, P, and K, respectively), a context for the scenario, and the internal state that the designed expressions should convey. 
\modified{The participants could design as many scenarios as time permitted. After the completion of each design, they were given a new scenario in randomized order.}

\subsubsection{Study Procedure} 
At the beginning of each design session, the experimenter offered an explanation of the study's purpose, such as ``exploring design possibilities with the use of the fluid expressions for robots,'' and provided instructions on how to use fluid expression devices and the robot. 

\begin{table*}[!t]
    \caption{Participant backgrounds, demographics, and familiarity with relevant media}
    \Description[Table]{The participants' background information is shown in the table with nine rows and seven columns.}
    \centering
    \small
    \label{table:background}
    \begin{tabular}{rp{0.18\linewidth}p{0.18\linewidth}p{0.18\linewidth}p{0.18\linewidth}ll}
    \toprule
    \textbf{ID} & \textbf{Background} & \textbf{Familiarity with animated films/anime} & \textbf{Familiarity with video games}  & \textbf{Familiarity with comics/manga} & \textbf{Age} & \textbf{Gender} \\ \toprule
    P1 & Education art                                                          & Regularly (daily/weekly)     & Occasionally (every few months) & Regularly (daily/weekly)     & 24                       & Male   \\ \midrule
    P2 & 2D design                                                              & Occasionally (every few months) & Occasionally (every few months) & Occasionally (every few months) & 20                       & Female \\ \midrule
    P3 & 3D art, ceramics                                                       & Regularly (daily/weekly)     & Occasionally (every few months) & Rarely                          & 20                       & Female \\ \midrule
    P4 & Graphic design                                                         & Occasionally (every few months) & Regularly (daily/weekly)     & Occasionally (every few months) & 21                       & Female \\ \midrule
    P5 & Graphic design                                                         & Occasionally (every few months) & Occasionally (every few months) & Occasionally (every few months) & 33                       & Female \\ \midrule
    P6 & Graphic design, computer animation (2D \& 3D), art metal, \& drawing & Regularly (daily/weekly)     & Regularly (daily/weekly)     & Regularly (daily/weekly)     & 22                       & Female \\ \midrule
    P7 & Graphic design                                                         & Regularly (daily/weekly)     & Rarely                          & Occasionally (every few months) & 20                       & Female \\ \midrule
    P8 & Character design, comic art                                        & Regularly (daily/weekly)     & Regularly (daily/weekly)     & Regularly (daily/weekly)     & 19                       & Female \\ \bottomrule
    \end{tabular}
\end{table*}

Following the introduction, participants engaged in a design activity that lasted 45 minutes. In the activity, the participant was given text-based scenarios that depicted a communication need for a robot and was asked to devise and demonstrate ways in which the robot expressed itself through the visual effects of fog, bubbles, and/or water droplets. Participants were asked to animate the robot using the control interface and add fluid expressions using the buttons connected to the expression devices. We did not impose any limitations on how to design and use the device in interaction design to not limit or bias participants' ideas. In addition to designing expressions, participants were asked to use a tablet computer to record their demonstration for each scenario. A semi-structured interview followed the completion of each design, then we provided them with another scenario to design.

Following the 45 minutes design activity, we conducted post-design session interviews that asked participants to reflect on their experiences during the session. The interview took approximately 10 minutes and was followed by a brief demographic questionnaire. 

All study activities were reviewed and approved by the University of Wisconsin--Madison Institutional Review Board (IRB).

\subsection{Participants}
Eight participants (1 male, 7 females), aged 19--33 ($M = 20.50$, $SD = 4.56$), participated in the design session. \modified{Because the design sessions aimed to explore the potential uses of fluid expressions in human-robot interaction design, we recruited participants with backgrounds in art and design. We expected this population to be able to integrate our design tools into a creative process and provide a design rationale. We recruited our participants via the mailing list of our campus Art Department. All participants were geographically located in the Madison, Wisconsin area and were fluent English speakers.
}
Table~\ref{table:background} provides a summary of participant backgrounds, demographics, and familiarity with relevant media (\textit{e.g.}, animated film, video games, comics). In the following sections, we refer to these individuals as \textit{P1}--\textit{P8}. Participants received \$15.00 USD as compensation.

\subsection{Measurement \& Data Analysis}
We collected two forms of data, including the video that captured designs from the participants and recordings from the interviews. The recordings from the interviews were transcribed into text.
To analyze our data, we followed a \textit{thematic analysis} approach \cite{thematicanalysis} by coding expressions designed by participants in the videos through annotation, transcriptions of justifications, and interview transcripts. We then categorized the codes to identify emerging themes and looked for examples from our data that illustrate each theme for presentation. 

We present our analysis from the design sessions in the next two sections. First, we describe the themes that emerged from our analysis of the designs and the justifications they provided in \S\ref{sec:findings:design}. Next, we describe the themes that emerged from our analysis of the interviews after the design sessions in \S\ref{sec:findings:interview}.


\subsection{Findings from Designs and Justifications} \label{sec:findings:design}
\begin{figure*}[!t]
        \includegraphics[width=\linewidth]{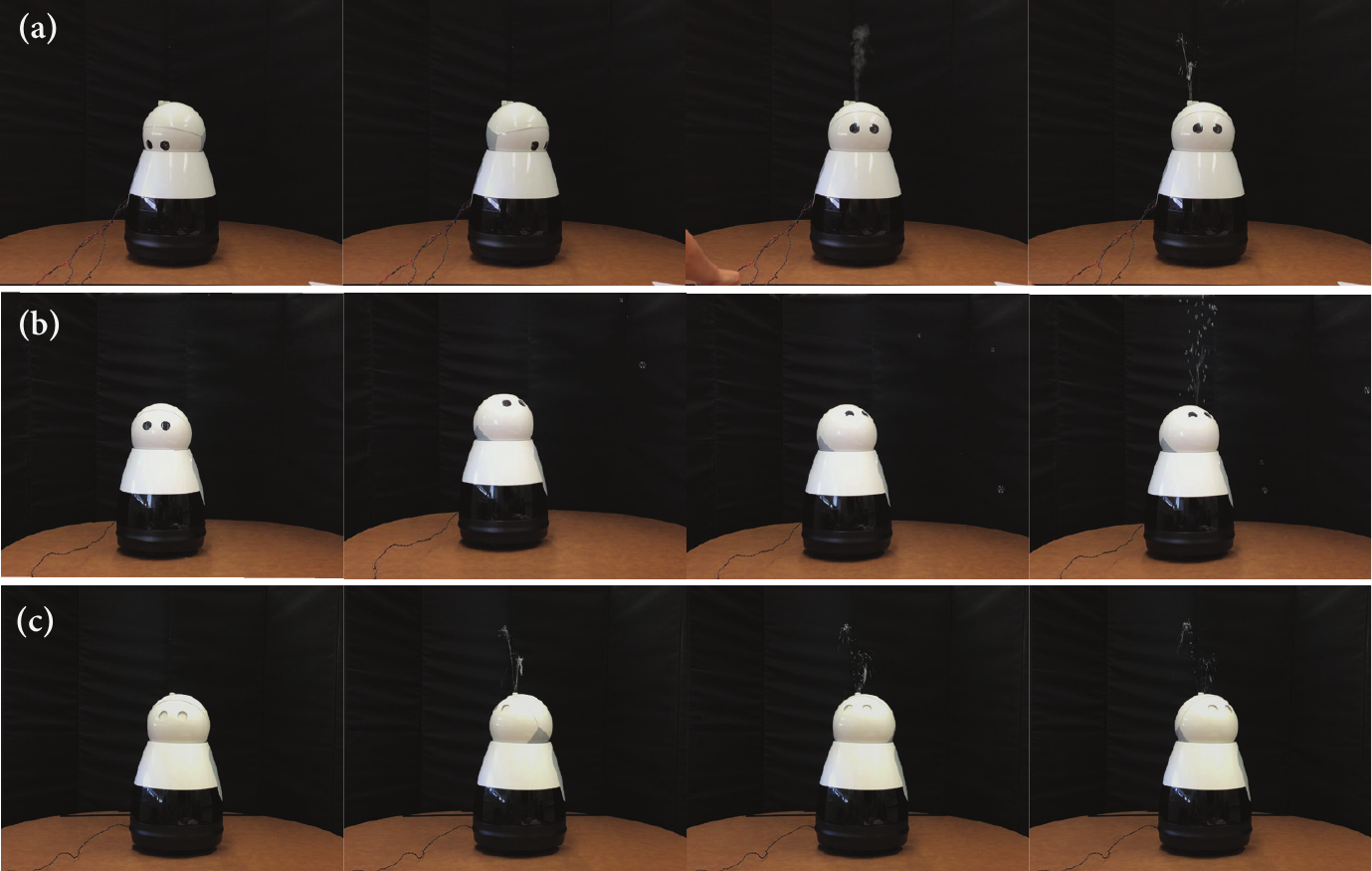}
        \caption{Snapshots from designs generated during the design sessions: (a) the robot looks around, first shows frustration with fog, and then expresses panic with water droplets (P1, Scenario D: \textit{delivery}); (b) the robot looks up, smiles with its eyes, and expresses curiosity using bubbles (P4, Scenario E: \textit{education assistance}); (c) the robot closes its eyes and sheds tears through water droplets while shaking its head (P8, Scenario K: \textit{playing with children}).}
        \Description[Screen Captures from the design session results]{Three final designs from the design session are labeled (a), (b), and (c). Each shows 4 keyframes from the final video from the participants. In (a), the robot first looks down-left, then down-right, and in the last two frames the robot looks straight and expresses fog generation and water droplet generation. In (b) the robot looks straight, then up-right then smiles and the last frame shows a robot with bubble generation. In (c), the robot looks straight, then in the remaining frames, the robot generates water droplets while looking around.}
        \label{fig:finaldesign}
    \end{figure*}

    From eight design sessions, we obtained 22 designs for five scenarios. Figure~\ref{fig:finaldesign} shows the excerpts of the designs from participants.
    Table~\ref{table:result} provides a summary of the designs from the participants.  Scenes from the design sessions are shown in Figure \ref{fig:setting} (f)--(h). Our analysis of the finals designs, design justifications, and interviews identified three main themes: 
    (1) Fluid expressions were used to convey specific robots' internal states; (2) How fluid expressions were combined with other non-verbal gestures; and (3) Participants found inspiration from media and stereotypes when designing robot expressions.
    Within each theme, we include quotes from the participant interviews, which are attributed using participant IDs in the interview transcripts (\textit{e.g.}, P2 denotes ``participant 2'').

\begin{table*}[!t]
    \centering
    \caption{Summary of which fluid mechanisms were used by each designer across scenarios.}
    \Description[Table]{The result of the design session (what kind of fluid was used in each scenario) is shown in the table with nine rows and six columns}
    \label{table:result}
    \begin{tabular}{cccccc}
    \toprule
       & \textbf{D (Frustration)} & \textbf{E (Curiosity)} & \textbf{P (Emergency)} & \textbf{G (Processing)} & \textbf{K (Feeling Hurt)} \\ \toprule
    \textit{P1} & Smoke, Water   & Bubbles       &               &                &               \\ \midrule
    \textit{P2} &                 & Bubbles       & Water         &                & Water         \\ \midrule
    \textit{P3} &                 &               &               & Smoke          &               \\ \midrule
    \textit{P4} & Smoke           & Bubbles       & Water         &                &          \\ \midrule
    \textit{P5} &                 &               &               & Water, Bubble & Water,  Smoke               \\ \midrule
    \textit{P6} & Smoke           & Bubbles       & Smoke         & Smoke          &               \\ \midrule
    \textit{P7} &                 & Bubbles       & Smoke         & Smoke          &               \\ \midrule
    \textit{P8} & Smoke           & Bubbles      & Water         &                & Water         \\ \bottomrule
    \end{tabular}
\end{table*}

\subsubsection{Theme 1: Fluid expressions were used to convey specific internal states of the robot}
    Our analysis of the designs showed that designers used specific types of fluids to convey specific internal states of the robot. At a high level, participants tended to use fog to convey frustration, bubbles to reflect positive context, and water to communicate unpleasant situations. We provide examples of designs and design justifications for each fluid type below.
        
    \paragraph{Fog} The fog was primarily used to convey frustration (Scenario D) and thinking (Scenario G). All participants who designed robot expressions for scenario D used continuous fog to express the robot's frustration and discussed their inspiration from cartoon expressions:
        \begin{quote}
            \textit{ ... Steam is like a common way to show \textbf{someone's really angry like cartoon stuff}.}  (P8)        
        \end{quote}
    Three out of the four participants who designed expressions for Scenario G used fog to express that the robot was thinking. P3 suggested that the fog around the robot's head served as a metaphor for intense thinking:
        \begin{quote}
            \textit{...I associate things or fog with like \textbf{fusion}, kind of like \textbf{brainstorming}, like \textbf{gears are turning}...} (P3)
        \end{quote}
    
    \paragraph{Bubbles} Bubbles were primarily used as an expression of curiosity (Scenario E). All participants who designed expressions for Scenario E used bubbles to express curiosity and mentioned the perceived positivity of bubbles in their design justifications. Moreover, five of the six participants combined the robot's smiley face with bubble generation. Three of the six participants described their motivation to use bubbles in terms of designing a positive expression for the robot:
        \begin{quote}
            \textit{When I saw bubbles, I kind of thought about `thought bubbles,' and like kind of generating thoughts and \textbf{obviously very happy}.} (P4)
        \end{quote}
    In addition, P2 and P8 referred to the unique positive characteristics of the bubble, compared to the other available fluid expressions:
        \begin{quote}
            \textit{I choose bubbles because it's \textbf{fluffy}, at most, I think, among those three elements.} (P2)
        \end{quote}
        \begin{quote}
            \textit{..Children will be interested because it's kind of like a \textbf{fun and interactive type of fluid}, and it would show the robot's happiness.} (P8)    
        \end{quote}
    
    \paragraph{Water Droplets} Five of the eight designs included the use of water droplets to express an unpleasant state. Three participants who designed expressions for Scenario K all used water droplets and stated that they intended for water to express crying. P3, in particular, described water droplets in terms of the robot's tears:
        \begin{quote}
            \textit{...I like water expression. It's the one most that's \textbf{similar to tears} ...} (P3)
        \end{quote}
    Three of the five participants (P1, P4 and P8) used water droplets for Scenario D, P or G described the use of water droplets as a reaction to an unexpected situation for the robot, such as sweat breaking due to stress or strain. P8 indicated that water droplets served as the robot's sweat to show that the robot was panicking:
        \begin{quote}
            \textit{... and uses like the two water spurts, because they're supposed to be like \textbf{sweat}, or something of a way of showing that it's kind of like \textbf{panicked}, and it needs to go over there right away ...} (P8)
        \end{quote}
   
\subsubsection{Theme 2: How fluid expressions were combined with other non-verbal gestures}
    Designers combined fluid with other behaviors such as head and body motion, gaze and eye expressions, rhythm, and repetition to create expressions that would follow familiar social norms and would provide social context for fluid expressions. For example, consider a scenario where the robot sprays water with no other body motion and another where the robot turns its head in a particular direction and sprays water. The head motion used in the latter scenario suggests that the fluid expression was in response to what the robot saw, while the use of fluid in the former is more ambiguous. We provide examples of how designers combined body movements and fluid expressions in unique ways.
    
    \paragraph{Head \& Body Motion} Combining fluids with head or body motions was used by participants to provide nuance and make expressions more explicit. For example, even though the robot moves at a fixed speed, P6 added fog to the robot when it is moving forward in order to convey urgency through faster movement:
        \begin{quote}
             \textit{... It just has one speed, but if it's emitting fog, then it kind of looks like, I guess, makes it \textbf{look like it's moving urgently}.} (P6)
        \end{quote}    
    P5 stated that the limited capabilities of the robot's head and body motions also limited the robot's expressivity, and such limitations create a need to over-exaggerate expressions:
        \begin{quote}
            \textit{Well, I definitely kind of over-exaggerated things a bit, because \textbf{it's a robot, and it's hard to get the emotions in the robot}. Probably if I was in the situation, I wouldn't be crying to the level of this robot or things, but to kind of show that.} (P5)
        \end{quote}
        
    \paragraph{Relationship with Eye Expressions} \label{sec:designsession:result:T2:eye}Participants frequently combined gaze with fluid expression to convey different emotions. For example, P4 combined blink with water droplets to show that the robot was surprised, while several participants used the combination of the smiley face and bubbles to express curiosity in Scenario E. P5 had the robot close its eyes when expressing breathing and waking up with bubbles for Scenario G. In the interviews for four of the five scenarios, participants discussed that more eye expressions of the robot would be helpful to realize their design ideas. Two participants told us about wanting the robot to squint its eyes along with fluid expressions. For example, P8 mentioned:
        \begin{quote}
           \textit{If it was possible to \textbf{make the eyes maybe squinted} like this, that we show \textbf{anger more}, ...} (P8) 
        \end{quote}
    In addition to squinting of the eyes, several participants expressed a desire for \textit{emoji}-like eye expressions. For example, P8 suggested that the use of greater-than and less-than signs (\textit{i.e.}, \texttt{>.<}) in the robot's eyes could express anger, panic, or excitement depending on the combination with fluids.
    
    \paragraph{Rhythm \& Repetition}         
    Some participants introduced rhythm and repetition to their use of fluid to create nuanced expressions. For example, P3 generated water droplets rhythmically to demonstrate child-like crying:
        \begin{quote}
            \textit{..when a baby cries and like in cartoons like they're like `wah wah wah', and it's kind of like it's like `one, two, three; one, two, three,' in this case, I did like \textbf{short, short, long}. So it looked like it was like `wah wah wah'...} (P3)
        \end{quote}
        

\subsubsection{Theme 3: Participants found inspiration from media and stereotypes when designing robot’s expressions}
    We asked participants to justify their design after each finalized design. These design justifications revealed that, during ideation, participants went through a sensemaking process of what each fluid expression meant and conveyed. In this process, they relied on different schemas, including prior exposure to popular media and stereotypes that linked fluids with people, events, and situations. 
    
    \paragraph{Participants found inspiration from media} Seven of the eight participants mentioned cartoons as a source of inspiration, and several participants specifically mentioned thought bubbles, which in part come from comic media. Overall, in six of the ten designs, participants took inspiration from cartoons and/or thought bubbles. Additionally, as discussed in Theme 1, all participants who designed expressions to convey frustration were inspired by prior depictions of steam coming out from the character's head. Several participants wished to have another nozzle for the fog machine. P8 stated:
        \begin{quote}
            \textit{.. if there was anything it would be steam go out both directions then it would show it's like \textbf{steam coming out from both ears that is the most popular expression for anger.}} (P8)        
        \end{quote}
    P1 stated that specific cartoons were sources of inspiration:
        \begin{quote}
            \textit{I guess, like \textbf{Looney Toons} and \textbf{Tom and Jerry} but those like really exaggerated expressions.} (P1)        
        \end{quote}  
    
    \paragraph{Participants relied on stereotypes} We saw that participants' reliance on stereotypes affected the expressions they selected. Especially for scenario E, all participants mentioned that ``children like bubbles'' when we asked them to describe their thought process in choosing to use bubbles. P7 and P8 stated:
        \begin{quote}
            \textit{Bubbles are \textbf{the most childlike and fun and playful}, I mean, and I think kids are more drawn to bubbles than they would be to water or steam.} (P7)
        \end{quote}
        \begin{quote}
            \textit{..Children will be interested [in bubbles] because it's kind of like \textbf{a fun and interactive type of fluid}, and it would show the robot's happiness.} (P8)    
        \end{quote}
    Similarly, P7 connected the fog expressions to ``sending smoke signals,'' stating:
        \begin{quote}
            \textit{..I want him to look around to assess the situation. And once you determine that there was something to \textbf{notify} whoever to \textbf{smoke signal to get attention}.} (P7)
        \end{quote}
    

\subsection{Findings from Post-Session Interviews} \label{sec:findings:interview}
    We also analyzed transcripts from post-design-session interviews that asked participants about their experience with the design session, including their general thoughts on fluid expressions and designing robot expressions with fluid expressions, which we report below.
    
    
    \subsubsection{Participants found designing fluid expressions challenging but fun}
    All participants described their experience with using fluid expressions in positive terms. For example, P3, P4, and P8 expressed that they \textit{``enjoyed''} it and that it was \textit{``super fun.''} P6 expressed that it was nice to \textit{`` be able to be a bit creative with design.''}
    Although some participants found it challenging to \textit{``use the liquids''} (P8), \textit{``figure out the logistics''} (P6), and  \textit{``work with fluids''} (P4), they also found the design tools we provided to be useful, stating that \textit{e.g.}, \textit{``timeline on the whiteboard was helpful''} (P1) and \textit{``being able to plan out what steps you want to follow''} were helpful (P7).

    \subsubsection{Design activities facilitated design suggestions from participants} 
    Most participants made design suggestions for other scenarios where fluid expression can be used or other expressions to make fluid expressions more effective.


        \paragraph{Alternative Scenarios} Five participants provided us with additional scenarios where the fluid expressions could be used, including using fluid expressions when interacting with children who \textit{``have a larger imagination''} (P1); using bubbles to indicate that the robot was ``sleeping'' or \textit{``minding its own business''} (P3);
        using specific forms of fluid to depict specific emotional states (\textit{e.g.}, \textit{bubbles} to show happiness, joy, excitement, and surprise; \textit{water} to convey surprise or mixed sadness; and \textit{smoke} to show anger (P7); and using fluids to convey robots' physical condition (\textit{e.g.}, smoke to depict that the robot is \textit{``damaged, that needs maintenance,''} (P6) or bubbles to express that the robot is swimming (P8).
        
        \paragraph{Alternative or Additional Expressions} Five participants (P2, P3, P4, P7, P8) desired more expressivity in Kuri's eyes, such as the addition of eyelids or eyebrows, to enrich and clarify the robot's internal states. As discussed in \S\ref{sec:designsession:result:T2:eye}, participants incorporated fluids and eye expressions to convey schemes of \textit{emoji} in their designs. 
        P2 and P7 recommended the additional use of color. Especially, P7 suggested using \textit{``different colored fluid,''} because people already have associations between colors and emotions. For example, red could be angry or blue bubbles could be happy.
        Finally, P1 suggested the use of airflow to express a sigh and lightning to express that the robot is shocked.
        
    \subsubsection{Participants shared their perceptions towards using fluid expressions for a robot}
    Participants (P1, P5, P6) enjoyed imagining how these expressions would be actuated on a real-life Kuri robot. For example, P1 explained, \textit{``I really like to imagine this actually being used, how people might react to those expressions''} and P5 expressed, \textit{``Well, I haven't done anything like this before, and I mean Kuri-chan is very cute!''}

    In the interview, we also asked which fluid expression(s) they liked. Nearly all participants liked bubbles because bubbles \textit{``make people happy to see that''} (P1) and \textit{``the bubble is kind of fun to see''} (P5). P4 liked the use of bubbles for situations such as scenario K because they \textit{``liked how the bubbles kind of invoked like this happiness that came with the curiosity of learning for that specific scenario.''}
    The fog was the second-most-popular fluid among the fluid expressions we provided. P1 and P4 liked how fog has associations with cartoon-like expressions. P1 expressed, \textit{``I think the steam is cool, it's kind of funny, so it's more of like the cartoon-like expression,''} and P4 expressed, \textit{''It's a staple of certain cartoons.''} P3 and P6 also liked fog because fog can be used widely; \textit{``the fog was most useful for [...] conveying like a wider range of expression in the robot''} (P6).
    Many participants liked water droplets because they can be used to express crying (tears) or sweat. However, one participant, P7, shared their concerns towards the water feature, expressing that the robot should be \textit{``water-proof''} and scenarios that use the water expression should be used when \textit{``people would already be prepared to get wet, maybe like in an outdoor situation, or ... aquarium''} to avoid having \textit{``wet floors.''}

\section{Discussion}

    Our design process and findings from the design sessions highlighted important design implications and limitations, which we discuss below in \S \ref{sec:discussion:implications} and \S \ref{sec:discussion:limitations}, respectively. Finally, the insights from the design process informed a promising design solution, \textit{EmoPack}, an attachment for robots in the form of a ``backpack'' that emits fluid-based cues to convey the robot's internal states and emotions, which we present in \S \ref{sec:discussion:emopack}.

\subsection{Design Implications}
\label{sec:discussion:implications}

We present six implications for design based on our findings from the design sessions and contextualize them in related work.


\subsubsection{\modified{Animation \& comics can be a resource for novel extra-linguistic cues}}


    \modified{In a future where we expect users of social robots to engage with them in complex ``relational'' ways \cite{turkle2006encounters}, we anticipate a need to use different types of cues for different communication needs. In focused, face-to-face, and conversational interactions, we might rely on expressions that are inspired by human expressions of emotion, such as facial expressions, as users might expect the cues they experience in human-human interactions to carry over to human-robot interactions \cite{breazeal2004designing}. In scenarios where implicit communication is more appropriate, such as the robot communicating its internal states as it performs the tasks it is designed to perform, the expressions might use cues from other sources, such as animation \cite{Takayama2011Pixar,szafir2014communication}. Designers might also identify scenarios where human-like expressions are combined with those inspired by animation and comics \cite[\textit{e.g.},][]{Young2007Cartooning,trovato2013cross}.}

    \modified{There are challenges, however, in designing expressive cues inspired by animation. Previous explorations of animation-inspired expression have relied on high-frame-rate animation techniques used in movies, such as those produced by Pixar \cite{Takayama2011Pixar} and Disney animation \cite{szafir2014communication}. These techniques require specific bodily capabilities or low-level manipulations of the robot's motion, which may not be feasible for many robots and robot design scenarios. In contrast, our work, relying on graphical tropes often used in low-frame-rate animations or static media such as comics, highlights an alternative source of inspiration for designing expressions for social robots.}

\subsubsection{\modified{Fluids can offer a promising design space for extra-linguistic cues}}

    \modified{
    We translated our inspiration from graphical tropes into cues that can be expressed through physical fluids and explored how designers might use them to design robot expressions. Our findings indicate that designers used a number of sources including commonly held stereotypes, prior exposure to popular media, and the match between biological processes and physical properties of the fluids to project meaning to fluid expressions. 
    Although we observed individual differences in interpretation, there was a noticeable agreement on the choice of fluids to convey information on specific states for the robot.  
    These observations expand existing literature \cite[\textit{e.g.},][]{Guo2020, Lee2019} on how fluids might be used in robot behavior design. Future research into this space can explore other materials, including wind, fire, lightning, and dust. For instance, fire could be used to express anger or jealousy, wind to express a deep sigh, and lightning to express shock or extreme surprise.}
 
    \modified{We also highlight that the addition of fluids to increase the robot's expressivity also extends the robot's \textit{materiality}, transforming the conventional robotic system that is experientially made up of solid materials to one that uses fluids in different forms. A growing body of literature explores robot materiality and its relationship to user experience, particularly with respect to the use of soft materials \cite[\textit{e.g.},][]{Hu2020,jorgensen2019constructing,jorgensen2022soft}. Our work points to fluids in different forms as a design material for expression as well as other potential uses.}


\subsubsection{Multimodal and multimedia expressions afford rich communication}

    Our analysis of the designs generated by our participants showed that \modified{using fluid expressions in combination with nonverbal cues resulted in richer expressions} and elevated their effect. The interviews with designers also revealed a desire to integrate additional modalities and media into the expressions, including light, color, and audio. For example, fog coming out from the robot's body might result in different expressions when red or white colors are projected onto the fog. 
    
    \modified{These design ideas point to a potentially new paradigm for human-robot interaction where, similar to 4DX theater, well-crafted multimodal expressions can create an immersive experience for the user. In a 4DX theater, various sensory effects such as vibration, scent, mist, wind, temperature changes, and strobe lights are used to enrich the movie experience for the audience. Similarly, in human-robot interaction, the use of multiple sensory modalities can create a more engaging and interactive experience between humans and robots. Prior work has also highlighted the promise of a multimodal approach in enhancing robot interactivity and creating evocative expressions, for example, by combining haptic feedback and visual information \cite{Mullen2021, Hu2020} and adding scent to a sweating robot \cite{Guo2020}. }

    

\subsubsection{Added expression can augment robot capabilities} 

    Our observations from the design sessions indicate that Kuri's behavioral capabilities were a significant limiting factor in creating highly expressive behaviors. 
    However, the addition of capabilities for fluid expression expanded the space of cues available for communication. Similarly, augmentations can focus on providing capabilities that robots might lack, such as the example of using AR to project arms on a robot that lacks arms \cite{Groechel2019}. 
    Such augmentations can be designed around supporting certain activities. For example, a device designed to transform an existing robot into a storytelling robot might enable the robot to use fog, stardust, projected light displays, and projections of additional features onto the robot that enables the robot to assume new characters, \textit{e.g.}, projection of wings to appear as an angel. Such augmentations can enable fantastical interactions with robots and substantially enrich human-robot interaction.
    

\subsubsection{Building and testing functional devices reveal design challenges}

    Although much of our exploration could be carried out in simulation or through scenario-based design, building the devices for fluid expression and giving them into the hands of designers have provided a wealth of insight into their potential use and practical challenges. We found, for example, that generating physical expressions was a fun and engaging activity for designers that provided them with a greater understanding of the expressive capabilities of fluids. We also learned the challenges associated with working with fluids. Water or moisture exposure can harm the robot's hardware or the environment within which the robot is used (\textit{e.g.}, hardwood floors, electronic equipment). Such challenges can be addressed through improved design; for example, the size of water droplets can be adjusted by designing the nozzle mechanism of the device, or the amount of moisture generated can be continuously measured and regulated by the device.
    

\subsubsection{Support for creative activity is critical in expression design}

    We found that participants utilized the design support tools and materials we provided, including the whiteboard and magnets for keyframing; supplies such as paper, pencil, and the robot figurine provided for ideation; and easy-to-use tools to generate robot behaviors and fluid expressions. These tools must support exploration and play in order to test and understand the boundaries of each expressive capability and minimize the technical knowledge required to control devices and systems. \modified{The literature on design education suggests that structured activities and tools can facilitate design creativity \cite{kowaltowski2010methods,mose2017understanding}. Future extensions of our work could include structured design activities and tools that complement our prototypes to facilitate design ideation and implementation.}

\subsection{Limitations}
\label{sec:discussion:limitations}
Our work has a number of limitations that fall under three categories: (1) study design, (2) study population, (3) cross-cultural differences, and (4) design options. 

\begin{figure*}[!t]
    \includegraphics[width=\linewidth]{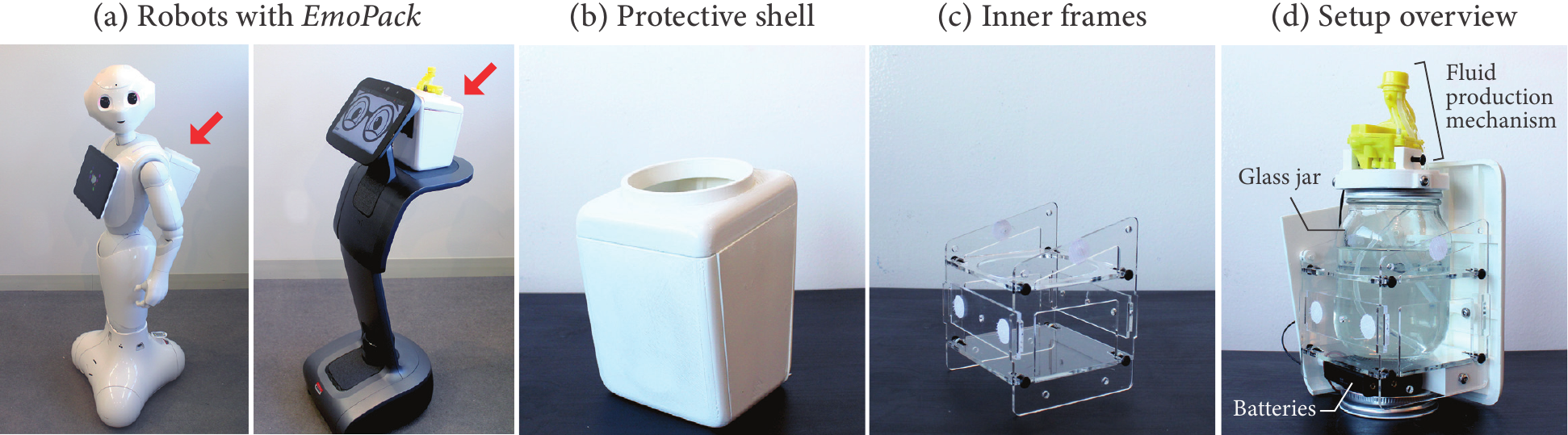}
    \caption{\textit{EmoPack} is an attachment for robots in the form of a backpack. We envision that EmoPack will be attached to (a) various robots to enhance their expressiveness (\textit{e.g.}, the Pepper and Temi robots). EmoPack consists of (b) a removable protective shell, (c) an inner frame to secure fluid expression mechanisms, and (d) the mechanisms themselves. }
    \Description[Annotated Photos]{Five photos inform the final design, called EmoPack, which is a backpack-style attachment for robots. The first and second photos from the left show that two kinds of robots have the EmoPack on their back. The third to fifth photos from the left show the detailed configuration of the EmoPack. The third photo shows the protective shell, the fourth photo shows the inner frames and the last one shows the setup overview.}
    \label{fig:EmoPack}
\end{figure*}

\subsubsection{Study Design}
    A drawback of the design session was the limited time allocated, with each session lasting only 45 minutes, which resulted in participants exploring a limited set of scenarios. This decision was made to avoid participant fatigue but also imposed some constraints on the scope of their creativity. Additionally, the scenarios presented to the participants during the session may not have been comprehensive, potentially limiting their ability to generate diverse designs. Several participants offered alternative design suggestions and scenarios in post-design interviews, indicating that there may be more opportunities to apply fluid expressions in robots. The scenarios presented could have also biased the design outcome in unexpected ways. For instance, while participants used bubbles to convey curiosity, they also expressed concern that children would favor them over other fluid expressions, which may lead to an overuse of bubbles as a design element if the intended audience was children. In future studies, we plan to promote more open-ended tasks without time restrictions to foster greater exploration and creativity.
    
\subsubsection{Study Population}
\label{sec:discussion:limitations:studypopulation}
    The results of the design session may be limited by the small sample size and specific background of the participants. Since most of our participants were young and familiar with popular media, their backgrounds may have influenced the final designs they created, which could affect the generalizability of our findings. In future work, our observations can be enriched by the participation of professional animators, puppeteers, or film directors. In addition to having designers' perspectives, viewers' perspectives can also be considered in future exploration. This includes investigating whether individuals without a design background or exposure to cartoons are able to effectively understand the internal states and emotions being conveyed by fluid expressions. Additionally, exploring how age, social factors, and cultural background can affect perception can help us better understand the nuances and variations in how fluid expressions are perceived by different populations. 
    

\modified{
    \subsubsection{Cross-cultural Differences}\label{sec:discussion:limitations:cultural}
    Graphical tropes originate in specific cultural contexts, and more research is needed to understand cultural differences in how extra-linguistic cues designed with inspiration from these cultural artifacts are perceived by people. Our analysis identified tropes that were used across cultural contexts, for example, characters leaving a trail of dust appeared in both American cartoons (\textit{e.g.}, Looney Tunes \cite{youtube_looneytoones}) and Japanese comic books (\textit{e.g.}, Doraemon \cite{doraemon_1} and Bakabon \cite{bakabon_1}) as an expression of rapid movement. However, how these cues will be perceived across cultures is unknown. Prior human-robot interaction that explored cross-cultural perceptions of extra-linguistic cues projected on the head of a humanoid robot found significant differences in participant preferences for these cues across Western, Asian, and Egyptian participants \cite{trovato2013cross}. More broadly, cross-cultural research in emotion found significant differences in the cues that Western and Eastern participants rely on in emotion recognition. Specifically, Japanese participants rely more on contextual cues than Western participants do \cite{masuda2008placing}. Cross-cultural differences also appear in the use of the tropes based on embodied experiences of the culture, for example, a Swiss cartoon using an ``avalanche'' to communicate a natural disaster and a Thai cartoon using a ``tsunami'' \cite{forceville2016conceptual}. Therefore, we expect some cross-cultural differences in what tropes might be semantically appropriate and how these tropes might be perceived by people. More research is needed to understand the extent to which tropes can be universally used or designed for specific cultural contexts.
}

\subsubsection{Design Options}
    The mobility and design of our prototype may have limited the design activity. First, the use of a mobile shelf system for carrying fluid devices was meant to provide some flexibility to the participants and simulate a fluid expression system that was integrated with the robot, but it also created some restrictions. The shelf system required participants to physically move the devices, which affected the appearance of fluid expressions, especially when the robot moved. Furthermore, the mobile shelf system did not allow for the dynamic movement of the robot, potentially hindering the process of design generation. 
    
    Second, the design of the fluid devices was also a limiting factor in the design sessions. Our implementation of the fog generator had nozzles that pointed upwards and horizontally, but some participants expressed a desire for splitting the fog in two different directions to simulate steam coming from the robot's ears. The water droplet generator also had a fixed pattern and shape, which limited the participants' ability to customize the expressions. The ability to tailor the speed, repetition, and shape of the water droplets would enable designers to create different effects, such as sweat-breaking or tears of joy. Similarly, adjusting the parameters of bubbles would also enhance their expressiveness, such as representing sleepiness or daydreaming through sparse and slowly moving bubbles.

\subsection{Final Design of \textit{EmoPack}} \label{sec:discussion:emopack}                     
    For future work in this area, we propose \textit{EmoPack} as a promising improvement of the prototype we used in our design sessions.
    \textit{EmoPack} offers a way to add extra-linguistic cues on a robot with limited expressive capability. We present EmoPack 1.0 as the first prototype of a \textit{backpack-style} attachment for robots that emits fluids in various forms to express the robot's internal states. EmoPack 1.0 addresses a key limitation faced in our design sessions related to the limited mobility of fluid devices. With the backpack-style EmoPack, designers can now attach fluid expression devices directly to the robot, allowing for greater creative freedom and flexibility.
    We believe that EmoPack 1.0 opens up new possibilities for further design sessions and field studies with increased scope for creativity.
\subsubsection{System Configuration} 
    Figure \ref{fig:EmoPack} shows the hardware configuration of \textit{EmoPack}, which consists of three modules: (1) protective shells, (2) inner frames, and (3) fluid expression devices.
    
    \textit{Protective shells ---} The 3D-printed shells protect the inner modules from damage and enhance the appearance of the robot with a glossy white finish, but other colors and shapes can also be used. Our final design features a backpack style, but this is not the only possible form. It could take the form of a hat, handbag, or any other style of attachment that is feasible and appropriate for the context.
    
    \textit{Inner frames ---} The inner frames secure and support the fluid device. For our design, we utilized acrylic casting for the frames, however, alternative materials like aluminum or stainless steel sheets can also be considered depending on the requirements.
    
    \textit{Fluid expression device ---} The fluid device module consists of a glass jar that holds the liquid and the mechanisms for generating fluid expressions as shown in Figure \ref{fig:EmoPack}. The fluid generation mechanisms remain largely unchanged from the design used in our sessions, except for the housing of the bubble machine being updated to match the glass jar style of the other fluid devices. Currently, each fluid expression requires a separate fluid device module, but future versions of EmoPack could use a single fluid expression module for multiple expressions.
    The modular design of EmoPack 1.0 is intended to provide designers and developers with the flexibility to customize the system for use with any robot platform. We provide all resources to develop \textit{EmoPack} in our Open Science Foundation repository.

\subsubsection{Use Cases}
    We envision a variety of potential applications for the EmoPack. Firstly, we envision that \textit{robot manufacturers} might adopt its variations into their offerings, for example, as a complementary accessory, to elevate the robot's expressiveness for specific purposes, such as storytelling.
    Secondly, \textit{end users} can integrate EmoPack into their existing robots to customize them and enhance their expressive capabilities, such as a school looking to assist students in regulating their emotions and having the robot demonstrate these emotions for them. Lastly, \textit{researchers} can use EmoPack as a platform to continue exploring fluid expressions in robotics, such as self-driving cars and behavioral research.

\section{Conclusion}

In this paper, we explored the design space of \textit{extra-linguistic cues} for social robots by addressing the following research questions: \textit{what metaphors might enable the design of effective extra-linguistic cues for social robots}, \textit{how can we implement extra-linguistic cues for robots with limited expressive capability}, and \textit{how designers use extra-linguistic cues to design robot expressions}. 

We found inspiration from \textit{graphical tropes} that are widely used in animation, video games, and comic books to convey the character's intent and emotions. Through a screening process to determine which graphical tropes can be applied to robot expression, we identified three groups of graphical tropes that take the form of fluids: \textit{smoke/steam/fog}, \textit{water droplets}, and \textit{bubbles}. We explored how they can be transformed to physically generate \textit{fluid expressions} and prototyped three fluid-generation systems. Using these prototypes and a Kuri social robot, we conducted design sessions that explored how designers perceived and used fluid expressions to express a social robot's intent and state in a scenario-based design activity. Findings from the design sessions revealed how fluids can be used for robot expressions, including how each form of fluid can serve as a communicative cue in each scenario and how these cues can be combined with other robot behaviors, \textit{e.g.}, head motion and facial expressions that were conveyed by the shape of eyes in the Kuri robot. 
Our study illustrates the promise of new extra-linguistic cues for human-robot interaction design as a unique and appealing communication modality and points to future research directions, such as understanding the cross-cultural use and interpretations of extra-linguistic cues.

\begin{acks}
    \modified{We would like to thank Rei Tamaru for the development of the Kuri robot controller used in the design sessions and Bengisu Cagiltay, Laura Stegner, and Pragathi Praveena for their assistance in the ideation, data analysis, and the writing of the paper. This work was made possible by financial support by the \textit{Sheldon B. and Marianne S. Lubar Professorship}, the \textit{H.I. Romnes Faculty Fellowship} and \textit{Shigeta Education Foundation}, and the in-kind donation of the Kuri robot by \textit{Mayfield Robotics}.}
\end{acks}

\balance
\bibliographystyle{ACM-Reference-Format}
\bibliography{99_ref}

\end{document}